\documentclass[letterpaper, 10 pt, conference]{ieeeconf}  

\IEEEoverridecommandlockouts                              

\usepackage{graphics} 
\usepackage{epsfig} 
\usepackage{mathptmx} 
\usepackage{times} 
\usepackage{amsmath} 
\usepackage{amssymb}  
\usepackage{algorithm}
\usepackage{algpseudocode}
\usepackage{hyperref}
\usepackage{subcaption}
\usepackage{xcolor}
\usepackage{multirow}
\usepackage{balance}

\DeclareMathAlphabet{\mathcal}{OMS}{cmsy}{m}{n}
\usepackage[T1]{fontenc}

\title{\LARGE \bf
ColNav: Uncertainty-Aware Spacetime Collision Checking \\ with Bounding Volume Hierarchies}

\author{Sanjeev Ramkumar Sudha, Lars Ivar Hatledal, and Erlend M. Coates 
\thanks{The authors are with the Norwegian University of Science and Technology, 6009 Ålesund, Norway. Email:
{\{\tt\small sanjeev.k.r.sudha, laht, erlend.coates}\}\tt\small@ntnu.no}
}

\begin{document}

\maketitle
\thispagestyle{empty}
\pagestyle{empty}

\begin{abstract}
Safe navigation for mobile robots in dynamic and uncertain environments requires planning paths that avoid collisions with other moving targets. Typically this involves forward propagating the robot motion and checking for collisions at sampled points in space and time. Such methods can miss collisions between sampled points and are not robust to uncertainty in the state estimates.  
We propose ColNav, a robust collision check formulation with geometric volumes.
The key aspect of our approach is encoding the uncertainty in the targets' and the ego robot's positions and velocities into spacetime tubes, in two variants that consume bounded-set and Gaussian uncertainty descriptions, respectively. The tubes are organised in bounding volume hierarchies, so that a single tree traversal covers the entire horizon, rather than testing sampled time instants, and returns a robust geometric certificate together with an estimate of the time to collision.
Static and dynamic obstacles share the same representation.
Our approach reaches the goal with a $100\%$ success rate under prediction model mismatch, where the best baseline reaches $65\%$, and maintains up to three times the clearance in dense traffic.
We also validate real-time operation on a physical robot.
\end{abstract}

\section{Introduction}

Mobile robots such as unmanned aerial vehicles (UAVs), ground vehicles (UGVs), and autonomous underwater vehicles (AUVs) typically operate in uncertain, dynamic environments and require real-time decision-making for safe navigation. Planning collision-free paths is necessary for robots operating in such environments \cite{latombe1991robot}. In practice, the robot can only estimate the states of the other agents through noisy sensor measurements, and the future motion has to be predicted \cite{lefevre2014survey}. 
Planning safe paths with efficient collision checking in dynamic environments with uncertain state estimates remains a challenge.



In this paper, we address uncertainty-aware collision checking for real-time planning in dynamic environments. The uncertainty in the predicted target states grows with the look-ahead time, and a check against a single predicted nominal trajectory becomes progressively less reliable over the horizon. Robust and fast planning instead requires efficient collision checks that reason about the set of states the ego robot and the targets can potentially occupy over the horizon.


A common way to check for collisions in candidate paths for the robot is to forward propagate the motion of the body and test for intersections with obstacle boundaries for sampled points in space and time \cite{lavalle2006planning}. These methods depend on time discretisation and can miss collisions that occur between sampled points, and can become inefficient for dynamic environments. Spacetime and continuous time methods treat the time-parametrised path as a curve in 3D \cite{Erdmann1987, Schwesinger2015}. However, these methods generally plan on nominal predicted trajectories and therefore ignore uncertainties in the estimated and predicted states.

\begin{figure}[t]
\centering
\includegraphics[width=0.9\linewidth]{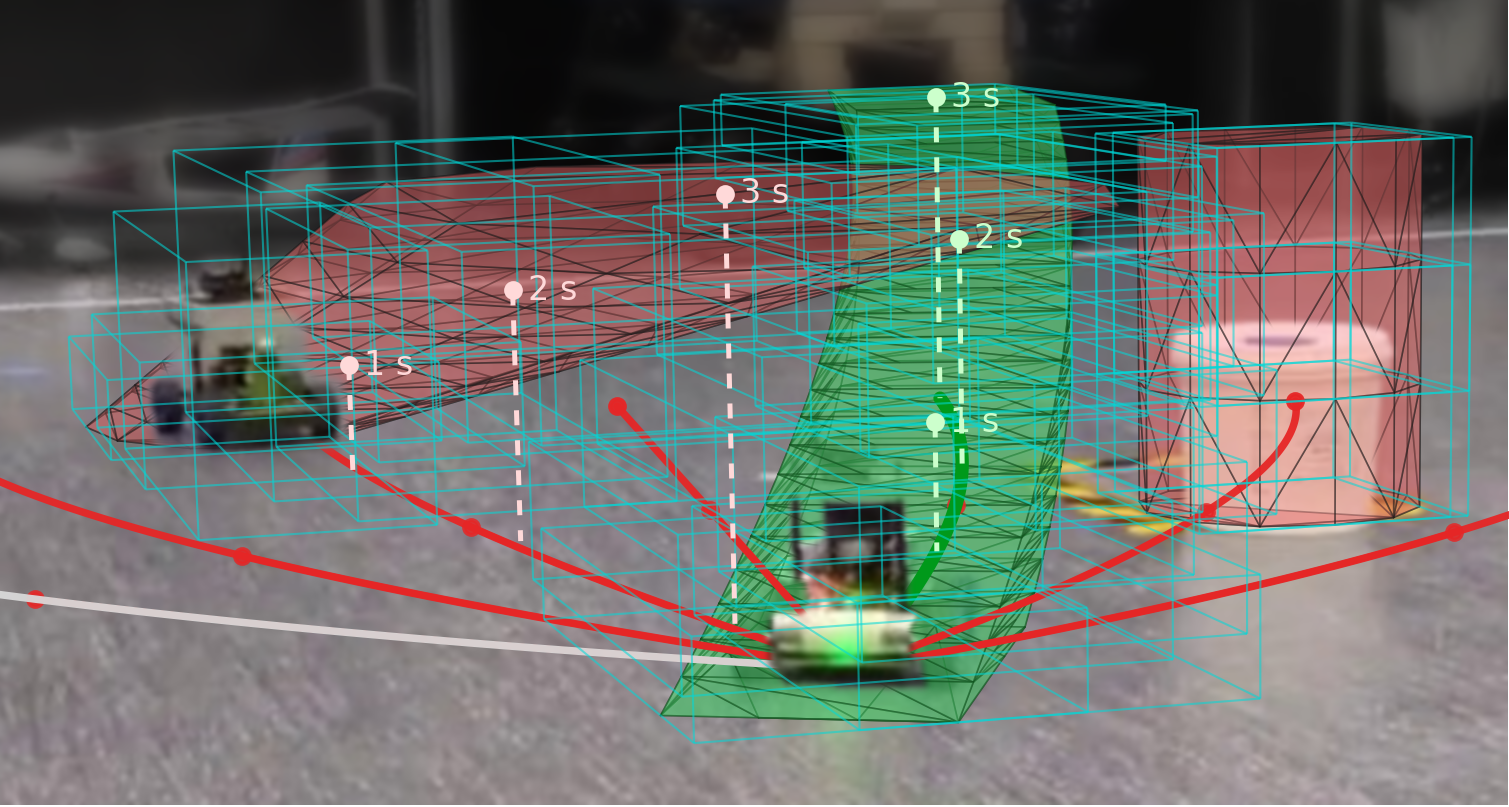}
\caption{Illustration of our approach in an experiment with UGVs. Our key idea is to turn uncertainty into geometry by encoding estimated and predicted uncertainty of each robot into its spacetime tube. A single BVH query over the whole horizon then certifies a trajectory and returns time to conflict.}
\label{fig:illus} 
\end{figure}


Localisation and target tracking systems typically provide covariance estimates or other uncertainty quantification of tracked targets' positions and velocities by fusing sensor observations \cite{bar2001estimation}. However, this information is rarely directly exploited in the planning or obstacle avoidance systems.
Probabilistic collision avoidance methods often evaluate a risk or probability of collision by integrating over the planning horizon \cite{du2012robot, patil2012estimating, mustafa2023probabilistic, blackmore2011chance}. They do not yield a fast worst-case geometric certificate to aid real-time planning requirements, for instance with sampling-based planners.

To address these challenges, we propose ColNav, a novel approach for robust spacetime collision checking under estimation and prediction uncertainty. ColNav formulates a worst-case collision check on bounded sets and Gaussian state uncertainty into geometric volumes. 
Our approach can support different types of uncertainty quantification and is a natural fit for both Gaussian filters and set-membership estimators. In this paper, we show uncertainties over the planning horizon can be propagated in the spacetime tubes using bounded sets and Kalman Filter predictions. For efficient volume representation and intersection checking, we use bounding volume hierarchies (BVHs). This enables efficient and robust collision checks that give a geometric worst-case certificate along with the time of collision, which can be used for real-time planning by integrating ColNav with sampling-based planners. We perform closed-loop evaluations for navigation with an object tracker and sampling-based planner, and validate real-time operation on a physical robot.
To summarise, our contributions are as follows:
\begin{itemize}    
\item A robust spacetime collision checking formulation that encodes estimation and prediction uncertainty in the tubes, with two variants for bounded-set (ColNav-R) and Gaussian uncertainty (ColNav-C).
\item We build spacetime tubes with BVHs and provide a unified representation for both static and dynamic obstacles. A single traversal of the tree returns a worst-case certificate under stated assumptions, along with the time to conflict.
\item 
Evaluations in simulation show fewer missed collisions and higher success rates than baselines that use constant uncertainty over the horizon, and a discrete checker with growing uncertainty bands.
Real robot experiments validate our approach for real-time planning. We will open-source our code to support further research. 
\end{itemize}


\section{Related Work}
\label{sec:related}

Configuration spacetime enables geometric collision checks for dynamic obstacles by extending the configuration space by appending time as an added dimension \cite{Erdmann1987}. Cameron \cite{Cameron1990} performed collision detection as a four-dimensional intersection test, extruding each body through time. For robot navigation in the 2D plane, operating in spacetime allows for utilising 3D graphics tools, enabling efficient 3D collision checks. An alternative approach is to characterise relative velocities of the target and ego robot that result in collisions \cite{Fiorini1998, van2011reciprocal}.
Related to our approach, Schwesinger et al. \cite{Schwesinger2015} used BVHs for efficient collision checking in spacetime. These works generally assume known motion and do not consider uncertainty while constructing the volumes. In contrast, we propose a robust collision checking method using BVHs that explicitly incorporates uncertainties from the estimated positions and velocities into the volumes. 

A BVH is a tree structure, typically binary, of bounding volumes, such as axis-aligned bounding boxes (AABBs), used for accelerated collision queries \cite{Gottschalk1996, ericson2004real}. Libraries such as the flexible collision library (FCL) \cite{pan2012fcl} support both discrete and continuous collision detection, and recent work accelerates both on GPUs using hardware ray tracing \cite{sui2025hardware}. Discrete queries can miss collisions between sampled time steps, and continuous queries cover a single motion interval per call. Although these accelerate repeated intersection tests, they operate on nominal geometry and assume perfect knowledge of the obstacles' motion. We build a BVH over each uncertainty-inflated spacetime tube. One traversal of a pair of BVHs covers the entire prediction horizon and returns time to collision from the earliest intersection. Every cross-section of the tube circumscribes the modelled uncertainty set, and both static and dynamic obstacles are handled with the same mesh representation and queries.

Probabilistic collision checking in uncertain dynamic environments has been studied using chance constraints \cite{du2012robot, mustafa2023probabilistic}. Collision risk has been studied assuming Gaussian noise and uncertainty \cite{patil2012estimating}, and also for non-Gaussian models \cite{park2020efficient}. These methods generally integrate collision probability or risk over time steps, but do not yield an efficient geometric worst-case certificate without time discretisation. 
In contrast to these approaches, we evaluate a robust collision test against a reachable set, which gives a worst-case certificate for a chosen confidence and uncertainty model.

In Hamilton-Jacobi reachability \cite{bansal2017hamilton} and set-based reachability analyses \cite{althoff2021set}, a set of reachable states is computed for verification and safe control. A body of work studies robust planning based on such approaches by bounding the ego robot's deviation under bounded disturbances \cite{gao2023collision, kousik2020bridging, herbert2017fastrack}. While these methods offer strong guarantees, they generally do not reason about the uncertainty of the other robots. Some works also reason about the uncertainty in the motion of other observed robots by growing the predicted region according to the estimated uncertainty \cite{fridovich2020confidence, althoff2016set}. Our approach is similar to this class of methods but differs in how the uncertain regions are represented. We formulate uncertainty as a spacetime tube from the target's predicted uncertainty and check for collision geometrically with a BVH, and a single query returns an estimate of the time to collision along with a worst-case certificate, suitable for integration with sampling-based planners.
%

\section{Problem Formulation}
\label{sec:formulation}

\subsection{Problem Statement}

We consider a robot (the \emph{ego robot}) navigating in 2D in an uncertain dynamic environment with static and dynamic obstacles (\emph{targets}). Given a nominal, planned trajectory $\gamma_o$ for some finite horizon $T$, we seek to certify if the trajectory is collision-free, taking into account uncertainty in localisation, control, and target motion predictions. 

In this work, we assume constant velocity (CV) dynamics for the targets. This could in principle be replaced with more elaborate trajectory prediction models for target intent. However, integrating our approach with such a system is out of scope of this paper.

\subsection{Workspace-time}

Each robot moves in the plane, with the position of robot $i$ at time $t$ denoted $p_i(t)\in\mathbb{R}^2$. We append time as a third coordinate to form the workspace-time~\cite{Schwesinger2015}, which we will refer to as just \emph{spacetime} for the remainder of this paper:
\begin{equation}
\mathcal{Z} = \mathbb{R}^2 \times \mathbb{R}_{\ge0}, \qquad z_i = (p_i,t).
\label{eq:spacetime_config}
\end{equation}

A time-parametrised trajectory is a curve $t\mapsto (p_i(t),t) \subset \mathcal{Z}$, i.e.\ the lift of the planar motion into spacetime. The key advantage of this lift is that the temporal coupling of a collision is encoded geometrically: two trajectory curves can only intersect at equal $t$, so a spacetime intersection automatically enforces simultaneity. Projecting a spacetime curve onto $\mathbb{R}^2$ recovers the geometric path. The 2-D path and the speed profile together determine the curve.

\subsection{From Curves to Volumes}

\begin{figure}[t]
\centering
\includegraphics[width=\columnwidth]{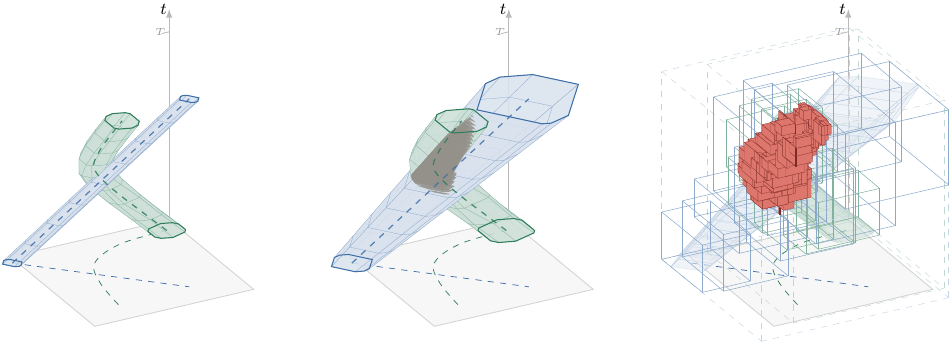}
\caption{Construction of the spacetime volumes for a crossing encounter between the ego robot (green) and a target (blue), over a horizon $T$. (left)~Sweeping the nominal trajectories (dashed) by the body extents, with the safety margin applied to the ego robot, gives tubes of constant cross-section. (middle)~Propagating the uncertainty gives $\mathcal{R}_o$ and $\mathcal{R}_i$, whose intersection is shaded. (right)~The same volumes organised in bounding volume hierarchies, with the conflict as the trees report it in red. The boxes over-approximate the intersection and serve only to prune. The triangle-triangle test at the leaves resolves it exactly for the
meshes.}
\label{fig:concept} 
\end{figure}

Robots have a finite physical extent. Robot $i$'s shape is modelled as a compact convex set $\mathcal K_i\subset\mathbb R^2$, where $\mathcal{K}_{i}$ is expressed in the robot's body frame, aligned to its heading $\psi_i$. 
We assume that the heading is aligned with the robot's velocity
direction (no drift), and that the body extent is aligned with the
nominal heading. The effect of heading errors on the body extent is
discussed in Sec.~\ref{subsec:test}. The shapes of objects being tracked are assumed to be known. However, this can be generalised by applying extended object tracking algorithms \cite{granstrom2016extended} to estimate the shape along with the kinematic states for objects of arbitrary shapes, but it is out of scope for this study.

For a finite time horizon, $t\in [0,T]$, the region occupied by robot $i$ in spacetime is a tube.
Given the spacetime tubes 
of two robots,
collision between the robots is a non-empty intersection of their tubes (\autoref{fig:concept}). A safety distance $d_s\ge0$ is incorporated by dilating the ego robot's tube by $B(d_s)$. 

\subsection{Uncertainty Representation}
\label{subsec:uncertainty}


To directly encode uncertainty into the spacetime tubes, we make appropriate assumptions and dilate the spacetime tubes, turning the intersection test into a robust collision check (see Fig.~\ref{fig:concept}).
The ego robot follows a nominal path $\gamma_o$ with speed $\bar u(t)$ through the means of a path-following controller and speed controller, respectively. Its localisation and tracking errors are small when compared to the uncertainty in targets' predicted states over the horizon. 
We therefore assume its spacetime volume does not grow over the considered horizon.
We consider the states of the targets to be estimated by the ego robot using a target tracking filter that estimates the pose and velocity of the targets. The target spacetime volumes grow throughout the horizon as the uncertain speed and velocity direction estimates are assumed to lie in some bounded sets. The specific volume obtained depends on our assumptions and how these sets are constructed, based on information provided by the tracking filter (Sec. \ref{sec:volumes}).

We write $\mathcal{X}_o(t)$ and $\mathcal{X}_i(t)$ for the resulting uncertainty
bands, the sets in which the position error of the ego robot $o$ and target $i$
are assumed to lie at time $t$. By the assumptions above, $\mathcal{X}_o$ is
constant over the horizon while $\mathcal{X}_i(t)$ grows with it, and both are
given explicitly in Sec.~\ref{sec:volumes}, where these bands are used to formulate the robust collision test.

\section{Spacetime Volume Generation for Robust Collision Checking}
\label{sec:volumes}
In this section we first state the robust collision test and the
assumptions under which it certifies a trajectory, and then present two
methods for constructing the uncertainty bands it relies on.
Fig.~\ref{fig:concept} illustrates the method. The tracker's
estimates and uncertainties are turned into uncertainty bands, either
from bounded sets (ColNav-R, Sec.~\ref{subsec:reachable}) or from EKF
predictions (ColNav-C, Sec.~\ref{sec:ekf_prediction}), and the resulting
volumes are meshed and organised in BVHs (Sec.~\ref{sec:impl_query}). One
traversal then evaluates the robust test of Sec.~\ref{subsec:test} over
the whole horizon and returns an estimate of the time to conflict.

\subsection{The Robust Collision Test}
\label{subsec:test}

In the robust formulation, every uncertain quantity is confined to a bounded set, and we determine whether a collision occurs for any admissible realisation.
A trajectory can collide with target $i$ if the spacetime reachable volumes intersect,
\begin{equation}
\mathcal{R}_o \cap \mathcal{R}_i \ne \emptyset ,
\label{eq:robust_test}    
\end{equation}
where $\mathcal{R}_o$ and $\mathcal{R}_i$ denote the reachable tubes of the ego robot and an arbitrary target $i$, defined as
\begin{align}
\mathcal{R}_o \,=\, \bigcup_{t \in [0,T]} \, \big(p_o(t) \,\oplus \, \mathcal{C}_o(t)\big)  \times \{t\}
\label{eq:reachable_tube_own} 
\\
\mathcal{R}_i \,= \,\bigcup_{t \in [0,T]} \,\big(p_i(t) \, \oplus \, \mathcal{C}_i(t)\big) \times \{t\}
\label{eq:reachable_tube_target_}
\end{align}
with cross-section of the tube at time $t$, $\mathcal C_i(t)$ computed as
\begin{equation}
\mathcal{C}_o(t)=\mathcal{K}_o\oplus\mathcal{X}_o\oplus B(d_s),
\qquad
\mathcal{C}_i(t)=\mathcal{K}_i\oplus\mathcal{X}_i(t),
\label{eq:crosssection}
\end{equation}
i.e., the body extent $\mathcal K_i$ inflated by the uncertainty band $\mathcal X_i $ and the safety margin $B(d_s)$ for the ego robot. A trajectory is declared collision-free only if~\eqref{eq:robust_test} fails for every target and every static obstacle. The resulting certificate holds under the following assumptions:
\begin{itemize}
  \item[(A1)] The heading error does not rotate the body extent, i.e.\
    $\mathcal{K}_i$ stays aligned with the nominal heading. A heading
    error $\theta$ moves a point of the body by at most
    $2r_i\sin(|\theta|/2)$, where $r_i$ is the circumradius of
    $\mathcal{K}_i$. This displacement can be absorbed by the safety
    margin $d_s$, as the test~\eqref{eq:robust_test} depends only on
    the combined margin of both robots.
  \item[(A2)] The ego tube, which follows a curved path, is resolved
    finely enough that the interpolation error between stations
    (Sec.~\ref{sec:impl_geometry}) is negligible compared to the safety
    margin~$d_s$.
\end{itemize}
If, in addition, the true states lie in the assumed bounded sets, a
negative test certifies that no admissible realisation collides over
the horizon. When the sets are confidence regions, the certificate is
conditional on the true states lying in those regions. For the
Gaussian regions of Sec.~\ref{sec:ekf_prediction}, this holds per time instant
under the linearised model, not jointly over the horizon.

The two methods below differ only in how the band $\mathcal{X}_i(t)$
in~\eqref{eq:crosssection} is obtained. The first (\emph{ColNav-R}) builds \emph{reachable tubes} by assuming the positions and velocities lie in bounded sets, giving a volume that contains every admissible realisation. The second (\emph{ColNav-C}) builds \emph{confidence tubes} by propagating the tracker covariance with an extended Kalman Filter (EKF) and thresholding at $k\sigma$, giving a volume whose cross-section at each time contains the target with a stated probability under the linearised prediction.




\subsection{ColNav-R: Spacetime Tubes from Bounded Sets}
\label{subsec:reachable}

In this section, we discuss one approach to compute the uncertainty bands, i.e.\ $\mathcal{X}(t)$ in \eqref{eq:crosssection} from bounded sets.

\emph{Targets:} A target $i$ at current time has estimated position $\hat p_i$, speed $\hat s_i$ and heading $\hat\psi_i$ with CV dynamics:
\begin{equation}
  p_i(t)=\hat p_i+t\,s_i\bigl(\cos\psi_i,\,\sin\psi_i\bigr)^{\!\top},
  \label{eq:cv}
\end{equation}
assuming the velocity is constant over the horizon but uncertain, with speed bands $s_i\in[\hat s_i-\delta s_i,\,\hat s_i+\delta s_i]$ and the heading band $\psi_i\in[\hat\psi_i-\delta\psi_i,\,\hat\psi_i+\delta\psi_i]$. Sweeping the bands, the displacement at time $t$ lies in the annular sector of radii $t(\hat s_i-\delta s_i)$ and $t(\hat s_i+\delta s_i)$ spanning $\pm\delta\psi_i$ about $\hat\psi_i$. We use the body-frame box, which is an outer approximation of the reachable set whose extents along the along-track and cross-track directions, $\hat u$ and $\hat n$, are given by:
\begin{align}
  \sigma_{\parallel,i}(t) &= k \sigma_{\parallel,i}(0)
    +\delta s_i t+\hat s_i t\bigl(1-\cos\delta\psi_i\bigr),
    \label{eq:along}\\
  \sigma_{\perp,i}(t) &= k \sigma_{\perp,i}(0)
    +(\hat s_i+\delta s_i)\,t\sin\delta\psi_i ,
    \label{eq:cross}
\end{align}
where $\hat s_i t(1-\cos\delta\psi_i)$ in \eqref{eq:along} accounts for the curvature of the sector, whose corners at $\psi_i=\hat\psi_i\pm\delta\psi_i$ fall short of the nominal displacement $\hat s_i t$. 
The uncertainty band $\mathcal{X}_i(t)$ is the body-frame box with half-widths $\sigma_{\parallel,i}(t)$ and $\sigma_{\perp,i}(t)$, which contains the sector for $\delta\psi_i \le \pi/2$ and $\delta s_i \le \hat{s}_i$.


\emph{Ego robot:} We assume the localisation error in the state estimation to be the main source of the position error, and take $\sigma_{\parallel,o}$, $\sigma_{\perp,o}$ to be constant for the ego. Therefore, the ego tube $\mathcal{R}_o$ has a fixed-size cross-section along the horizon that depends on the extent of the robot, the safety margin $d_s$ and a set from which we assume the initial position is contained.

The bands $\mathcal{X}_o$ and $\mathcal{X}_i(t)$ complete the cross-section \eqref{eq:crosssection},
and with it the volumes \eqref{eq:reachable_tube_own}-\eqref{eq:reachable_tube_target_}. Constant half-widths
$\sigma_{\parallel,o},\sigma_{\perp,o}$ are used for the ego robot, and the half-widths
\eqref{eq:along}-\eqref{eq:cross} for the targets.

\emph{Source of the bounds:} The construction is agnostic to how
the bounded sets are obtained. Set-membership estimators
under unknown-but-bounded noise, such as zonotopic Kalman
Filters~\cite{combastel2015}, return a set that provably
contains the state. For Gaussian filters, the bounded sets can be taken as $k\sigma$ regions of the estimated covariance and the reachable sets are conditional on the probability that the true state lies in those regions. In the latter case, we set $\delta s_i=k\sigma_{s,i}$ and $\delta\psi_i=k\sigma_{\psi,i}$, with the errors in heading $\sigma_{\psi}$, speed $\sigma_s$, and position typically obtained from an object tracker. If the covariances in the positions are known, 
the initial half-widths are $k \sigma_{\parallel,i}(0)$ and
$k \sigma_{\perp,i}(0)$, with the body-frame standard deviations computed as
\begin{equation}
  \begin{bmatrix} \sigma_{\parallel,i}^2 &
                  \sigma_{\perp,i}^2 \end{bmatrix}^{\!\top}
  = \operatorname{diag}\!\big( R_{\psi,i}^{\top} [P_i]_{pp}\,
    R_{\psi,i} \big),
  \label{eq:rot}
\end{equation}
where $[P_i]_{pp}$ denotes the covariance matrix of the position $\bar p=[x,y]^\top$, which can be obtained from the object tracker. The rotation matrix $R_{\psi,i}=[\hat u\;\;\hat n]$ transforms $[P_i]_{pp}$ into body frame, thereby obtaining $\sigma_{\parallel,i}$ and $\sigma_{\perp,i}$. For the ego robot we can use a $k\sigma$ envelope of its position covariance $[P_o]_{pp}$. 

\subsection{ColNav-C: Spacetime Tubes from EKF Predictions}
\label{sec:ekf_prediction}

\subsubsection*{Mean propagation with the tubes}
Each target is tracked by an EKF with state vector for target $i$ in the polar form $\mathbf{x}_i = [x_i,\,y_i,\,s_i,\,\psi_i]$, with CV dynamics described as follows: 
\begin{equation}
\label{eq:polar_cv_model}
\dot{\mathbf{x}}_i = [s_i\cos\psi_i \,\,\  s_i\sin\psi_i \,\,\ 0 \,\,\ 0]^\top,
\end{equation}
where target $i$ with an estimated position $p_i=[x_i,\,y_i]^\top$ moves at constant speed $s_i$ along constant heading $\psi_i$. Therefore, given states $\mathbf{x}_i$ of target $i$, the mean position at time $t$ is propagated with CV prediction \eqref{eq:cv}.

\subsubsection*{Covariance propagation with the tubes}
We propagate the EKF covariance forward in time with continuous-time EKF predictions:
\begin{equation}
  P_i(t) = \Phi_i(t)\,P_i(0)\,\Phi_i(t)^\top + Q_i(t),
  \label{eq:cov_prop}
\end{equation}
where $\Phi_i(t)$ is the linearised transition matrix, $\Phi_i = I + A_i t$, and $A_i$ is the Jacobian, and $Q_i(t)$ is the accumulated process noise over the horizon, from the injected acceleration and turn-rate noise. It is built from a continuous white-noise-acceleration (CWNA) model: on a single scalar axis $[p,\dot p]^\top$ driven by white acceleration of power spectral density (PSD) $q$, the accumulated covariance over a horizon $t$ is
\begin{equation}
  Q(t)= q\begin{bmatrix} \tfrac{t^3}{3} & \tfrac{t^2}{2} \\[2pt]
                          \tfrac{t^2}{2} & t \end{bmatrix}
\label{eq:cwna}
\end{equation}

For the polar CV model, white noise enters from acceleration (PSD $q_{s,i}=\sigma_{a,i}^2$) on speed $s$ and turn rate (PSD $q_{\psi,i}=\sigma_{\omega,i}^2$) on heading $\psi$, collected in $Q_{c,i}=\mathrm{diag}(q_{s,i},\,q_{\psi,i})$ and injected onto the velocity states through $G=[\,0_{2\times2}\ \ I_2\,]^\top$. Since the CV Jacobian is nilpotent ($A_i^2 = 0$), $\Phi_i(t) = I + A_i t$ is the exact transition matrix of the model linearised at the current estimate, and the accumulated process noise
\begin{equation}
  Q_i(t)=\int_0^t \Phi_i(\tau)\,G\,Q_{c,i}\,G^\top\,\Phi_i(\tau)^\top\,d\tau
\label{eq:Q_integral}
\end{equation}
evaluates in closed form to
\begin{equation}
  Q_i(t)=
  \begin{bmatrix}
    \tfrac{t^3}{3}\,J_i Q_{c,i} J_i^\top & \tfrac{t^2}{2}\,J_i Q_{c,i}\\[4pt]
    \tfrac{t^2}{2}\,Q_{c,i} J_i^\top      & t\,Q_{c,i}
  \end{bmatrix},
\label{eq:Q_full}
\end{equation}
where $J_i $ is the velocity Jacobian of the polar model and the non-zero block of $A_i$ \cite{schubert2008comparison}, defined as
\begin{equation}
J_i=\partial\dot p_i/\partial\nu_i=\begin{bmatrix}\cos\psi_i & -s_i\sin\psi_i\\ \sin\psi_i & s_i\cos\psi_i\end{bmatrix},
\label{eq:jacobian}
\end{equation}
where $\nu_i = (s_i, \psi_i)$. Mapping the velocity-state process noise into position through $J_i Q_{c,i} J_i^\top$ is the standard input-Jacobian construction of the state-space process noise \cite{bar2001estimation}.
Each block of \eqref{eq:Q_full} is the matching entry of the scalar CWNA structure \eqref{eq:cwna} with the scalar $q$ replaced by the two-channel $Q_{c,i}$.
Solving for $P_i(t)$ with \eqref{eq:cov_prop}, the position block of the predicted covariance, $[P_i(t)]_{pp}$, then evaluates to
\begin{equation}
\begin{aligned}
  [P_i(t)]_{pp} ={}& [P_i(0)]_{pp}
    + t\big( J_i [P_i(0)]_{\nu p} + [P_i(0)]_{p\nu} J_i^{\top} \big) \\
  &+ t^2 J_i [P_i(0)]_{\nu\nu} J_i^{\top}
    + \tfrac{t^3}{3} J_i Q_{c,i} J_i^{\top},
\end{aligned}
\label{eq:cov_pred_pos}
\end{equation}
and along and cross-track position errors $\sigma_{\parallel,i}(t)$, $\sigma_{\perp,i}(t)$ are obtained by projecting the predicted position covariance $[P_i(t)]_{pp}$ onto the body-frame axes. 


\emph{Anisotropic uncertainty ellipse:} unlike Sec.~\ref{subsec:reachable}, where
the band is the body-frame box \eqref{eq:along}-\eqref{eq:cross}, here
$\mathcal{X}_i(t)$ is the $k\sigma$ ellipse of the predicted position block
\eqref{eq:cov_pred_pos}, with body-frame semi-axes $k\sigma_{\parallel,i}(t)$ and
$k\sigma_{\perp,i}(t)$. The two bands enter the mesh construction of
Sec.~\ref{sec:impl_geometry} through their support functions alone, so the volumes,
the BVHs and the queries are otherwise identical.


\section{Numerical Implementation}
\label{sec:colnav}

In this section, we describe how the tubes \eqref{eq:reachable_tube_own}-\eqref{eq:reachable_tube_target_} are constructed as meshes and how the robust test \eqref{eq:robust_test} is evaluated efficiently with BVHs. Each volume is tessellated into a triangle mesh and organised in a BVH, a binary tree of nested AABBs. Two meshes are tested by descending both trees together while pruning every pair of subtrees whose boxes are disjoint \cite{Gottschalk1996, ericson2004real}. Both dynamic and static obstacles share a common representation. 
The meshes and BVHs of all targets and obstacles form a scene built once per control cycle. The BVH implementation is built using threepp \cite{threepp}, whose bindings provide the curve primitives and the BVH.

\subsection{Mesh Generation}
\label{sec:impl_geometry}

Each volume is built according to \eqref{eq:reachable_tube_own}-\eqref{eq:reachable_tube_target_}. The horizon is
divided into $n_t$ time stations, at each of which the cross-section
$\mathcal{C}(t)$ of \eqref{eq:crosssection} is approximated as a ring of $n_r$
vertices in the constant-time plane, and consecutive rings are joined by quads.
The construction relies on support functions. For a compact set
$\mathcal{A}\subset\mathbb{R}^2$ and a direction $d(\phi)=(\cos\phi,\sin\phi)^{\!\top}$, the support
function $h_{\mathcal{A}}(\phi)=\max_{a\in\mathcal{A}}d(\phi)^{\!\top}a$ locates the supporting line
of $\mathcal{A}$ with normal $d(\phi)$, so that $\mathcal{A}$ lies in the half-plane
$\{x:d(\phi)^{\!\top}x\le h_{\mathcal{A}}(\phi)\}$, and a convex set is the intersection of these
half-planes over all $\phi$~\cite{leguernic2010}. They are additive 
under Minkowski sums,
$h_{\mathcal{A}\oplus\mathcal{B}}=h_{\mathcal{A}}+h_{\mathcal{B}}$, and positively homogeneous,
$h_{\lambda\mathcal{A}}=\lambda h_{\mathcal{A}}$ for $\lambda\ge0$, which is what makes them
convenient here: a cross-section assembled from several uncertainty terms is evaluated as a sum of
scalars per direction, and no set is ever formed explicitly.

The support of~\eqref{eq:crosssection} is the sum of its terms,
\begin{align}
  h_{\mathcal{C}_o}(\phi) &= h_{\mathcal{K}_o}(\phi)
    + h_{\mathcal{X}_o}(\phi) + d_s \label{eq:support1}, \\
  h_{\mathcal{C}_i}(\phi,t)& = h_{\mathcal{K}_i}(\phi)
    + h_{\mathcal{X}_i}(\phi,t),
  \label{eq:support2}
\end{align}
where a rectangular body of length $l_i$ and width $w_i$ contributes
$h_{\mathcal{K}_i}(\phi)=\tfrac{l_i}{2}|\cos\phi|+\tfrac{w_i}{2}|\sin\phi|$ and the safety margin contributes the constant $d_s$ to the ego robot
only. Only the band term distinguishes the two
methods of Sec.~\ref{sec:volumes}:
\begin{align}
  h_{\mathcal{X}_i}(\phi,t) &= \sigma_{\parallel,i}(t)\,|\cos\phi|
    +\sigma_{\perp,i}(t)\,|\sin\phi| , \label{eq:hXbox}\\ 
  h_{\mathcal{X}_i}(\phi,t) &= k\,\sqrt{d(\phi)^{\!\top}\Sigma_i(t)\,d(\phi)} ,
    \label{eq:hXell}
\end{align}
where \eqref{eq:hXbox} is the box of ColNav-R, with half-widths
\eqref{eq:along}-\eqref{eq:cross}, and \eqref{eq:hXell} the ellipse of
ColNav-C, with $\Sigma_i(t)=R^{\!\top}_{\psi,i}[P_i(t)]_{pp}R_{\psi,i}$ the
predicted position block \eqref{eq:cov_pred_pos} rotated into the body frame. Both reduce
with $k=0$ to the body extent and the safety margin alone, and the ego robot is
obtained from either with constant half-widths.
 
The ring is the polygon of the supporting lines at the $n_r$ directions, each
vertex the intersection of two consecutive lines. A convex set lies inside all
of its supporting half-planes, so the ring circumscribes $\mathcal{C}(t)$ and
each ring is a certified outer approximation of its cross-section: the rectangular terms are reproduced exactly when the sampled directions include the body axes, since the supporting lines then pass through the corners, while the curved terms are over-covered by at most a factor $\sec(\pi/n_r)$.
 
Between stations the mesh interpolates linearly. For the targets and
static obstacles, this is exact whenever the centre moves at constant
velocity and the half-widths are affine in~$t$, as they are for ColNav-R
by~\eqref{eq:along}-\eqref{eq:cross}. For ColNav-C, the interpolation
over-covers wherever $h_{\mathcal{X}_i}(\phi,t)$ is convex in~$t$, as
it is absent a negative position-velocity correlation. For the ego
robot, whose path is curved and whose body axes rotate, the interpolated
mesh only approximates the swept volume. Its error per interval is at
most $\kappa_{\max}\ell^2/8 + r_o\,\Delta\psi$, where $\ell$ and
$\Delta\psi$ are the arc length and heading change between consecutive
stations, $\kappa_{\max}$ is the maximum path curvature and $r_o$ the
circumradius of $\mathcal{K}_o$. Assumption~(A2) requires this error to
be small compared with~$d_s$.
 
\emph{Targets:} the station centres lie on the segment from $(\hat p_i,0)$ to
$(\hat p_i+\hat s_i T\hat u_i,\,T)$ given by the CV prediction \eqref{eq:cv},
and the body axes are constant over the horizon.
 
\emph{Ego robot:} the waypoints, each with a leg speed $\bar u_j$, define a
Catmull-Rom path that is sampled at $n_s$ points equally spaced in arc length,
lifted to time by
\begin{equation}
  t_j = t_{j-1}+\|p_j-p_{j-1}\|/\bar u_j ,
  \label{eq:lift}
\end{equation}
and thinned to the $n_t$ stations with a central-difference tangent at each. By
Sec.~\ref{subsec:uncertainty} the cross-section is the same at every station, so
only the centre and the body axes vary along the horizon.
 
\emph{Static obstacles} are prisms spanning $[0,T]$: a cylinder for circular
regions, or a polygon for arbitrary, possibly concave, footprints.
A concave footprint is extruded directly rather than through \eqref{eq:support1}-\eqref{eq:support2},
which assumes a convex cross-section. The BVH and the queries of
Sec.~\ref{sec:impl_query} are unaffected.

\subsection{BVH Construction and Queries}
\label{sec:impl_query}

Each mesh is organised in a BVH by recursively sorting its triangles by centroid along the longest axis of the node box and splitting at the median, until a node holds at most $L$ triangles \cite{ericson2004real}.
The robust test is the simultaneous descent of both trees. Node pairs with disjoint boxes are pruned, and leaf pairs are resolved by Möller's exact triangle-triangle test \cite{Moller1997}, 
which also yields the segment on which two triangles intersect. These
segments sample $\partial\mathcal{R}_o \cap \partial\mathcal{R}_i$, as
shown in Fig.~\ref{fig:concept}. Before the traversal, the cross-sections at $t=0$, i.e.\ the first rings of both meshes, are tested for overlap in the plane, using a
separating-axis test for convex rings. If they overlap, a conflict with
$t^\star_i = 0$ is reported. Otherwise, the volumes intersect if and
only if their lateral surfaces do, since cross-sections that are
disjoint at $t=0$ must touch before they overlap. The time-to-conflict
$t^\star_i$ is estimated from the earliest segment midpoint. Each
triangle spans one station interval $\Delta t = T/(n_t-1)$, so this
estimate exceeds the true earliest conflict time by at most
$\Delta t/2$ (0.09\,s in our setup). The position coordinates of
the same segment locate the conflict.

\section{Results and Discussion}
\label{sec:experiments}

In this section, we evaluate ColNav and compare it with fast discrete and continuous collision checkers. First, we evaluate how reliably ColNav detects collisions by driving the robot on a fixed nominal path, and test its robustness in scenarios with prediction model mismatch and degraded sensors. In the closed-loop evaluation, each checker is used inside a sampling-based receding-horizon planner, and we measure success rates in reaching the goal and minimum clearance to obstacles.
Real robot validation is also performed with a UGV.

\subsection{Setup}
\label{sec:setup}


We use the F1-tenth simulator \cite{babu2020f1tenth} and use an Ackermann-steering UGV for the experiments. For tracking targets in the environment, the joint probabilistic data association (JPDA) filter \cite{bar2009probabilistic} is used. Both versions of ColNav are driven by the same JPDA tracker so that the comparison isolates the uncertainty envelope model rather than the estimator. For ColNav-R the initial sets are formed as $k\sigma$ regions of the tracker's position and velocity covariances. With ColNav-C, the covariance is propagated directly.

We evaluate ColNav-R and ColNav-C against the following collision checkers chosen as baselines: (i) a discrete checker from FCL where we inflate the targets by a $k\sigma$ envelope of $\sigma_{\parallel, i}(0), \sigma_{\perp,i}(0)$ constant over the horizon; (ii) FCL discrete checker with targets inflated with the bounded set uncertainty \eqref{eq:along}-\eqref{eq:cross}; (iii) a swept volume BVH checker \cite{redon2005fast}, analogous to the approach of \cite{Schwesinger2015}, with target uncertainty accounted the same way as (i); The extent of the ego robot is inflated with the safety margin and the localisation uncertainty for all checkers, as described in Sec. \ref{subsec:test}. 

Three scenarios are used for the evaluations, where the ego robot is moving on a nominal path towards a goal and has to avoid approaching targets: (i) scenario 1 with two targets crossing at $2-3$ times the speed of the ego; (ii) scenario 2 with two targets manoeuvring at a constant turn rate; (iii) a traffic scenario with five crossing targets with constant velocities. We simulate the scenarios under two sensor noise conditions, with $\eta=0.02$ m for the nominal lidar range noise, and $\eta=0.10$ represents a degraded sensor case.

For the closed-loop evaluation, we evaluate ColNav and the other collision checkers for real-time planning with a sampling-based receding horizon planner. With ColNav, the target and obstacle BVHs are built and checked for collisions against each candidate trajectory. The planner evaluates a finite set of discretised control behaviours with each candidate $\gamma_j$ corresponding to a fixed speed $u_j \in \{0,..,u_c\}$ m/s and heading $\psi_j \in \{-\psi_m,..,\psi_m\}$ over the planning horizon. Each candidate trajectory is given a cost:
\begin{equation}
  J(\gamma_j)\,=\,5\,c\left(1 - \frac{t^{\star}_j}{T}\right) \,+\, 2\,\frac{|u_j-u_c|}{u_c} \,+\, \frac{|\Delta\psi|}{\pi/2} \,+\, \frac{|\Delta u|}{u_c},
  \label{eq:planning_cost}
\end{equation}
where $c \in \{0,1\}$ and $t^{\star}_j$ is the minimum time to a conflict if there are collisions, and $c=0$ in case there are none. The costs penalise changes in speed $\Delta u$ and heading $\Delta \psi$, with an additional penalty slowing from the cruise speed $u_c=2$ m/s. We use a horizon $T=3$ s with time steps of $0.3$ s.


\subsection{Open-loop evaluation}

\begin{figure}[!htb]
    \centering
    \begin{subfigure}{0.23\textwidth}
        \centering
        \includegraphics[width=0.95\linewidth]{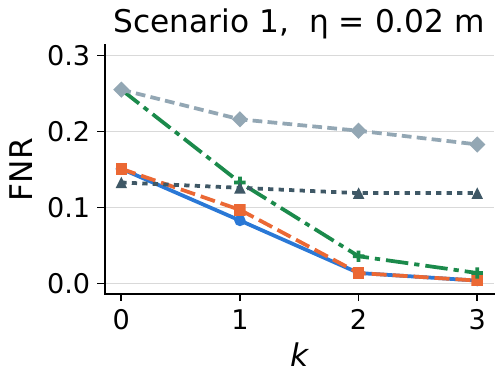}
    \end{subfigure}
    \begin{subfigure}{0.23\textwidth}
        \centering
        \includegraphics[width=0.95\linewidth]{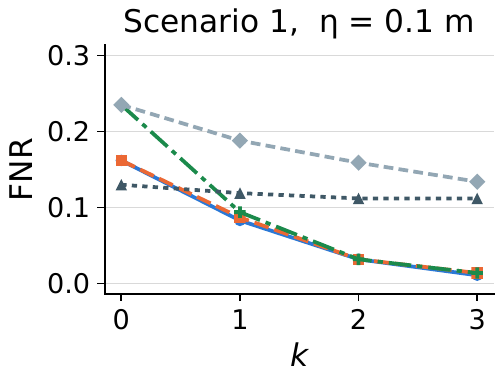}
    \end{subfigure}
    \begin{subfigure}{0.23\textwidth}
        \centering
        \includegraphics[width=0.95\linewidth]{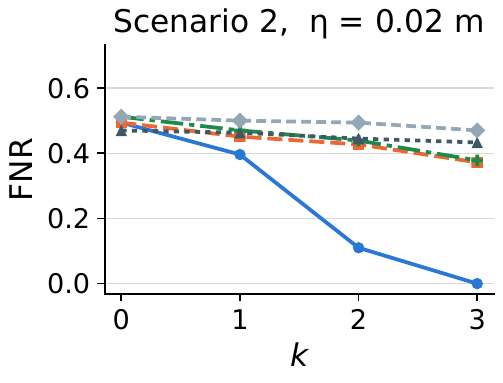}
    \end{subfigure}
    \begin{subfigure}{0.23\textwidth}
        \centering
        \includegraphics[width=0.95\linewidth]{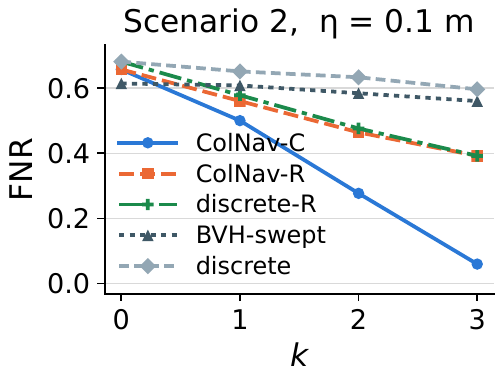}
    \end{subfigure}
    \caption{FNR plotted for scenarios $1$ and $2$ under nominal and degraded sensor conditions. Discrete and BVH-swept maintain a constant $k\sigma$ envelope over the horizon. Discrete-R uses the same growing bands as ColNav-R but tests at sampled times.
    }
    \label{fig:fnr_plots}
\end{figure}

In this experiment, we simulate the ego robot along a fixed nominal path. 
The collision detectors are evaluated on their false negative rates (FNR), computed by checking for collisions at the planning steps using the predicted motion of the targets with the CV model against ground truth by testing if bodies actually overlap during this horizon.
We run these evaluations over $20$ trials for $k=0,1,2,3$, and the cumulative FNRs are summarised in \autoref{fig:fnr_plots}.

\begin{table*}[!htb]
\centering
\begin{tabular}{|l|ccc|ccc|ccc|ccc|}
    \hline
     & \multicolumn{3}{c|}{Scenario 2 ($\eta=0.02$)}
    & \multicolumn{3}{c|}{Scenario 2 ($\eta=0.10$)}
    & \multicolumn{3}{c|}{Scenario 3 ($\eta=0.02$)}
    & \multicolumn{3}{c|}{Scenario 3 ($\eta=0.10$)} \\
    Strategy & Rate & $d_c$ & $t_g$
             & Rate & $d_c$ & $t_g$
             & Rate & $d_c$ & $t_g$
             & Rate & $d_c$ & $t_g$ \\
    \hline
            
    Discrete & $0\%$ & $0.00$ & - & $0\%$ & $0.00$ & - & $95\%$ & $0.13$ & $\mathbf{10.8}$ & $80\%$ & $0.20$ & $\mathbf{11.0}$ \\
    Discrete-R & $15\%$ & $0.01$ & $7.6$ & $30\%$ & $0.01$ & $7.7$ & $95\%$ & $0.32$ & $11.2$ & $\mathbf{100\%}$ & $0.48$ & $11.4$ \\
    BVH-swept & $65\%$ & $0.04$ & $7.5$ & $10\%$ & $0.00$ & $7.6$ & $\mathbf{100\%}$ & $0.31$ & $11.0$ & $95\%$ & $0.49$ & $11.3$ \\
    ColNav-R (k=2) & $55\%$ & $0.03$ & $7.5$ & $60\%$ & $0.05$ & $7.7$ & $95\%$ & $0.38$ & $11.4$ & $95\%$ & $0.52$ & $11.6$ \\
    ColNav-C (k=2) & $\mathbf{100\%}$ & $\mathbf{0.20}$ & $\mathbf{7.4}$ & $\mathbf{70\%}$ & $\mathbf{0.09}$ & $\mathbf{7.5}$ & $\mathbf{100\%}$ & $\mathbf{0.98}$ & $11.9$ & $\mathbf{100\%}$ & $\mathbf{0.78}$ & $11.8$ \\

    \hline
    \end{tabular}
    \caption{Closed-loop evaluations with a sampling-based planner.}
    \label{tab:results}
\end{table*}

In both scenarios, the FNR largely remains flat with increasing $k$ for the baselines that assume constant uncertainty throughout the horizon. In scenario 1, the FNR falls to near zero with both ColNav variants and discrete-R for $k=3$, whereas for the swept-BVH checker, it is $0.122$, despite the FPR being comparable for all the checkers, in the range of $0.05-0.07$. This reveals that an uncertainty envelope growing with time reduces missed collisions, as opposed to inflating the tubes with a constant $k\sigma$ envelope.
In scenario 2 with the manoeuvring targets, a new source of uncertainty enters, i.e., the error in the prediction model, and the gap between the baselines and ColNav-C is larger. ColNav-C performs better than ColNav-R in this case, even under sensor degradation.
This is attributed to ColNav-C accounting for the uncertainty in the prediction model through the process noise $Q_c$, whereas ColNav-R only propagates uncertainties in the estimated velocity. However, this conservative nature of ColNav-C also results in an increase in FPR at $k=2$ by up to $5$ times that of ColNav-R and up to $2$ times greater under sensor degradation.

\subsection{Evaluation with a sampling-based planner}
\label{sec:closedloop_results}

We perform closed-loop evaluations with the ego robot being controlled with the sampling-based planner, with planning costs as given in \eqref{eq:planning_cost}. Simulations are repeated for 20 trials with each checker at $k=2$, and we define success rate as $\%$ of trials where the ego robot avoids collisions and reaches the goal. The minimum clearance ($d_c$) is defined as the true distance between the bodies at the closest point of approach. The mean time to reach the goal ($t_g$) is computed only for successful runs. The results are reported in \autoref{tab:results}.

We observe that ColNav-C records the highest clearance and success rates in all scenarios. 
In scenario 3, discrete-R is comparable to ColNav-R on all metrics, but in scenario 2 with the manoeuvring targets, the rate of the former is lower at both noise levels. This indicates that despite using the same uncertainty bands, a discrete checker can fail under prediction uncertainty.
In scenario 2, the best-performing baseline only reaches the goal in $65\%$ and $30\%$ of the trials under the two sensor noise cases. ColNav-C reaches the goal in all trials with $\eta=0.02$ and in $70\%$ of trials with $\eta=0.1$, while ColNav-R reaches it in $55\%$ and $60\%$ of the trials.
The baselines reach the goal more reliably in scenario 3, but report lower clearances compared to the two ColNav variants.
In scenario 3, ColNav-C records $3$ and $1.6$ times the clearance of the strongest baseline, the swept BVH checker under the two noise conditions, with only $8.2\%$ and $4.6\%$ added time to reach goal.

The mean compute time for collision checking with ColNav with $5$ targets using $35$ candidate trajectories is $22$ ms with an Intel Core i7-12700H CPU. 
The discrete checker takes $2$ ms, but it operates on box primitives and does not traverse a tree. The swept BVH checker takes $42$ ms.

\subsection{Real Robot Experiments}

\begin{figure}[!htb]
    \centering
    \begin{subfigure}{0.23\textwidth}
        \centering
        \includegraphics[width=0.95\linewidth]{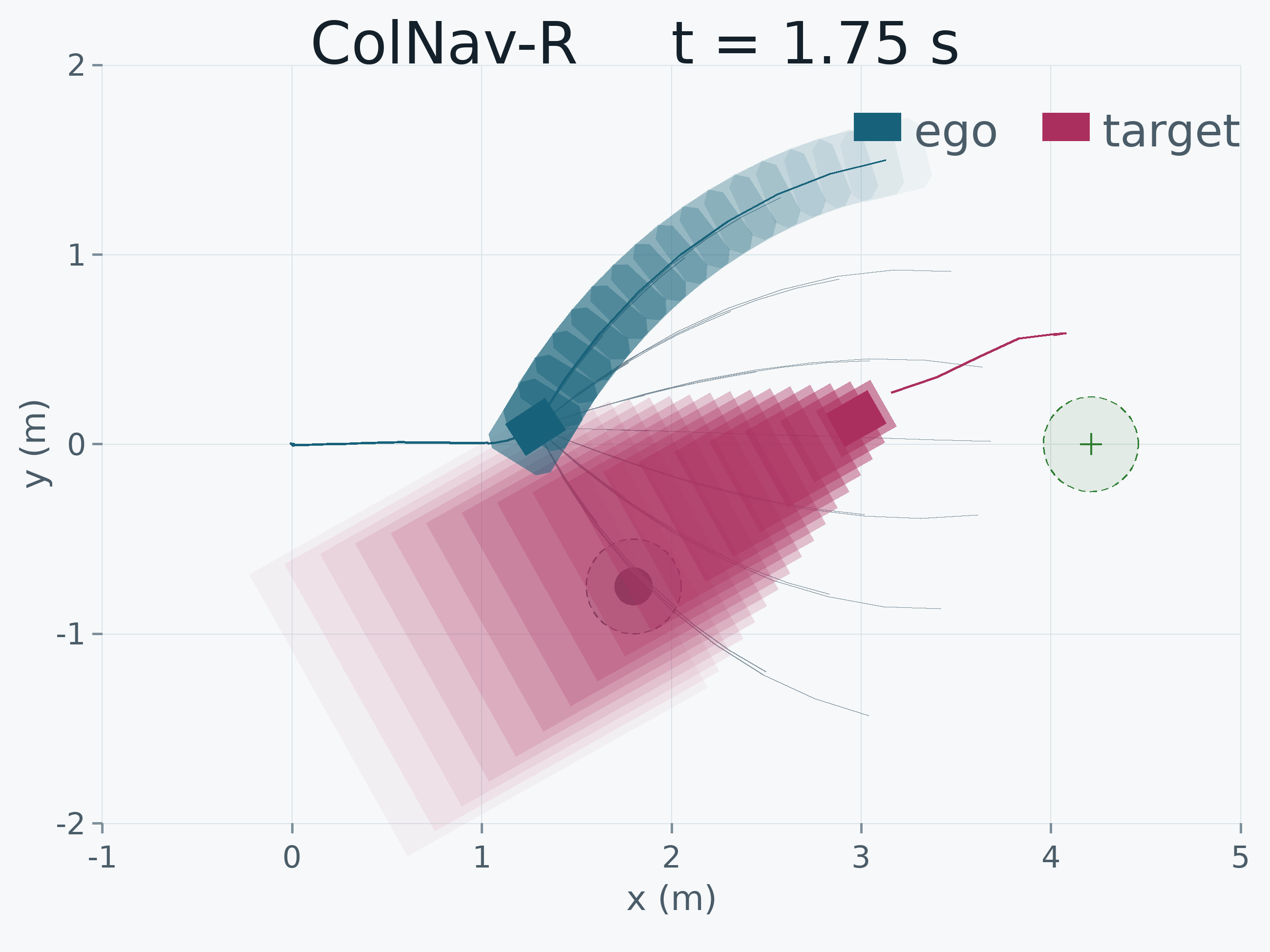}
    \end{subfigure}
    \begin{subfigure}{0.23\textwidth}
        \centering
        \includegraphics[width=0.95\linewidth]{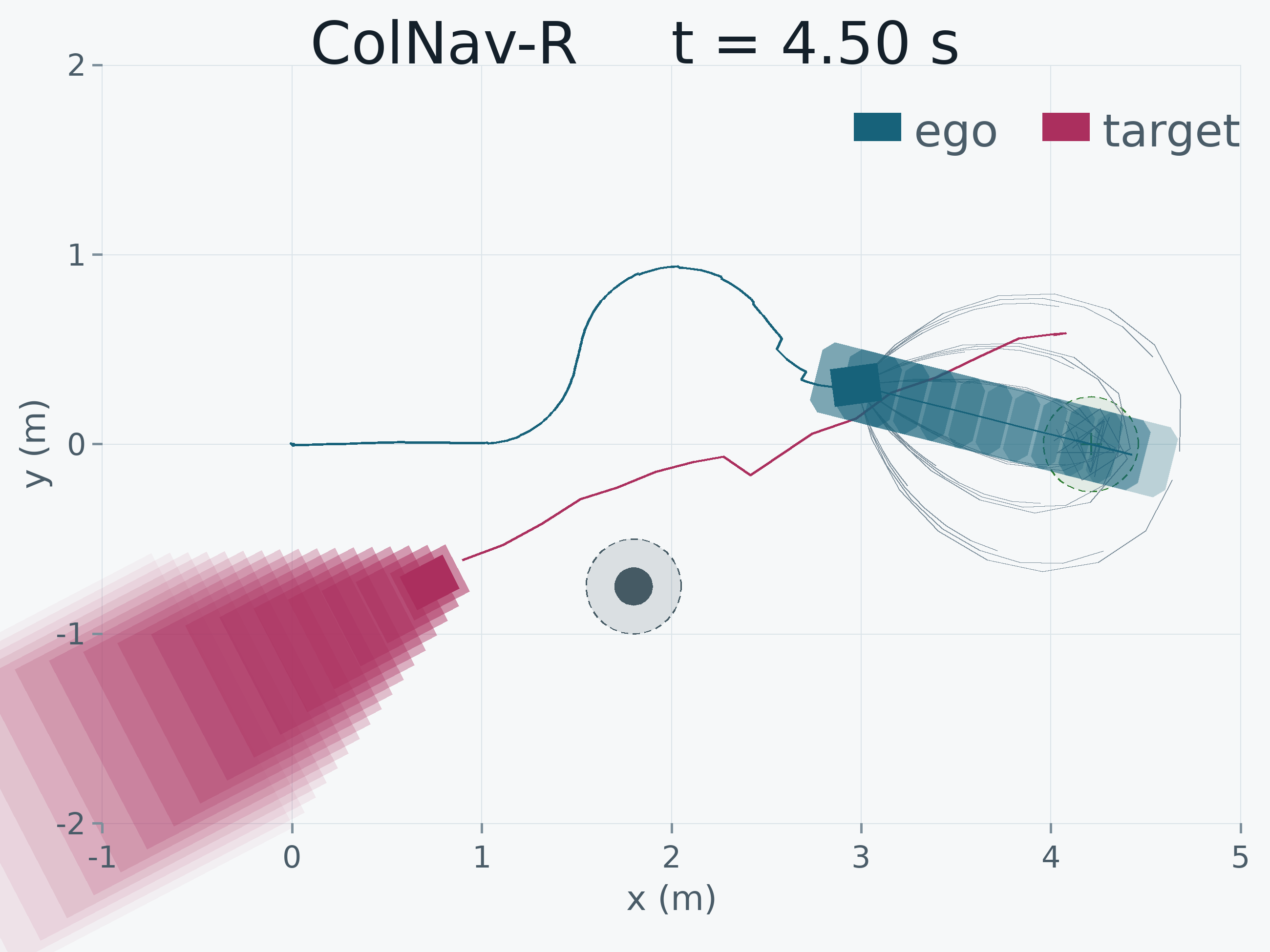}
    \end{subfigure}
    \\[0.35em]
    \begin{subfigure}{0.23\textwidth}
        \centering
        \includegraphics[width=0.95\linewidth]{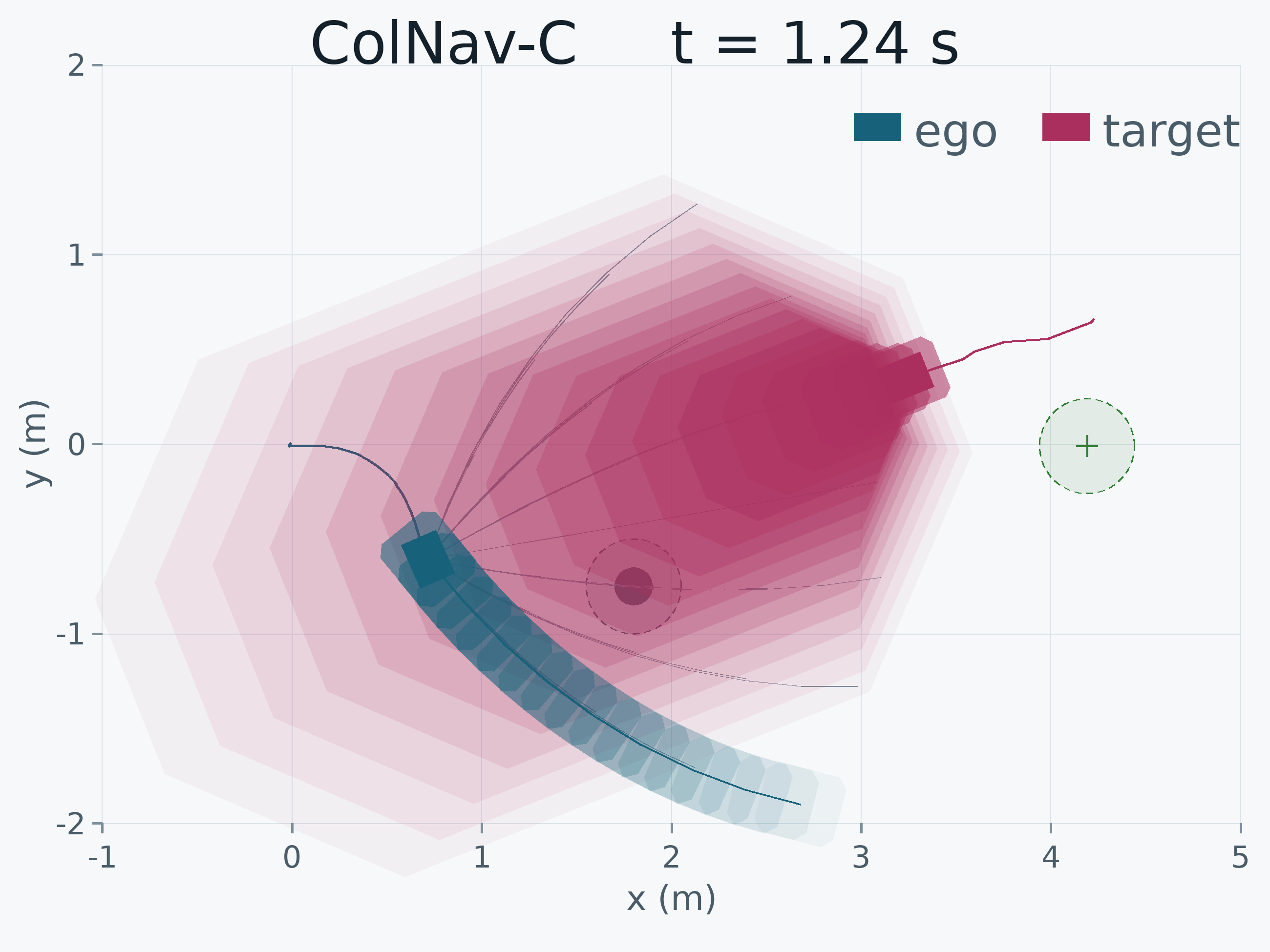}
    \end{subfigure}
    \begin{subfigure}{0.23\textwidth}
        \centering
        \includegraphics[width=0.95\linewidth]{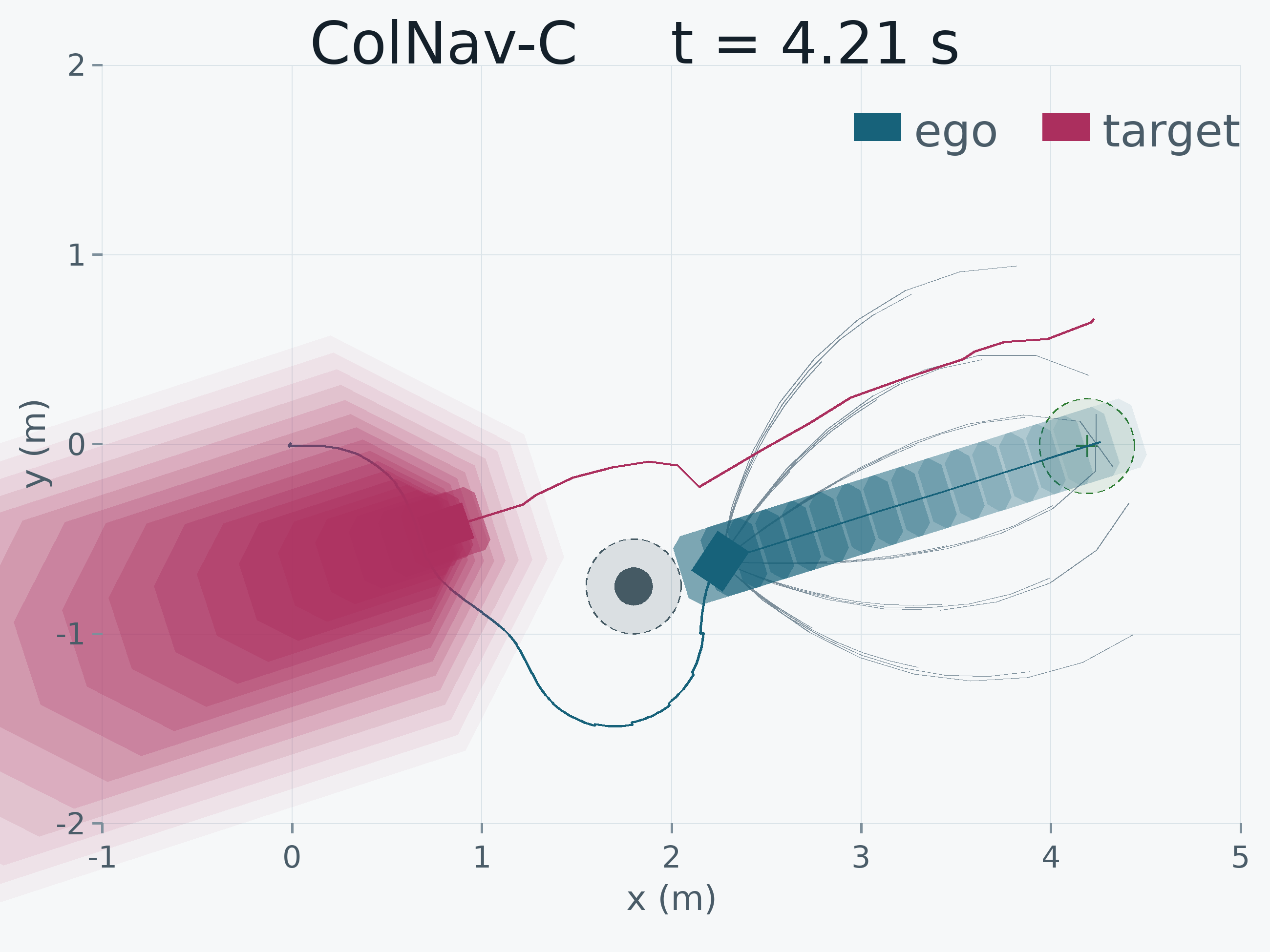}
    \end{subfigure}
    \caption{Snapshots from individual real robot trials depicting the obstacle, ego with its path followed, the target and its tracked trajectory, and a top-down view of their spacetime tubes. The planned path for the ego and the target predictions and their tubes are also visualised.}
    \label{fig:real_exp}
\end{figure}

We also validate our approach with end-to-end perception and collision avoidance planning on a Jetracer UGV (see \autoref{fig:illus}). We implement a JPDA filter to track targets and run the sampling-based planner at $ 3.3$ Hz using ColNav with $35$ candidate trajectories, on a Jetson Nano with $ 4$ GB RAM. The ego robot navigates to a goal $4.25$ m away with $u_c=1$ m/s, in an environment with one static obstacle and one target that moves with a mean speed of $0.8$ m/s. We repeat five trials each with the two ColNav methods, and summarise the results in \autoref{tab:realrobot_exp}. ColNav-C records a lower $d_c$ at the closest point of approach to an obstacle, contrary to the results in Sec. \ref{sec:closedloop_results}. This is because in the experimental scenario, the conservative nature of ColNav-C makes the ego robot choose the safer but longer path at the start of the trial that drives it around the static obstacle, as seen in \autoref{fig:real_exp}. It avoids the target early but records lower clearance as it drives closer to the static obstacle. Meanwhile, with ColNav-R, the ego robot drives straight toward the goal and then stops and takes a longer turn to avoid the incoming target.

\begin{table}[!htb]
    \centering
    \begin{tabular}{|c|c|c|}
    \hline
        & $d_c$ & $t_g$ \\
       \hline
       ColNav-C & $0.191 \pm 0.117$ & $6.47 \pm 0.36$ \\
         ColNav-R & $0.387 \pm 0.305$ & $6.71 \pm 0.82$ \\
    \hline
    \end{tabular}
    \caption{Mean $d_c$ and $t_g$ from the real robot experiments.}
    \label{tab:realrobot_exp}
\end{table}

\section{Conclusions and Future Work}

In this paper, we propose a novel approach for robust collision checking in spacetime. With our approach, prediction and estimation uncertainties are encoded into spacetime tubes, based on bounded sets and Gaussian filter uncertainty propagation.
It also provides a unified representation for static and dynamic obstacles, with both built as BVHs, and a single tree traversal gives a deterministic certificate and a time to collision. Evaluations show that our approach achieves greater success rates under degraded sensor conditions and prediction model mismatch as compared to baselines that maintain a constant uncertainty over the horizon, or use a discretised checker with the same uncertainty bands. ColNav-C reports near-zero FNR at $k=3$ under sensor degradation and a mismatched prediction model. We also validate our approach in real robot experiments with a UGV. 

This study focuses on worst-case collision testing using spacetime volumes.
It gives a worst-case certificate based on the geometry for a chosen confidence level or uncertainty model. The certificate relies on assumptions (A1)-(A2). Removing them, by adding a rotation margin and an ego interpolation margin to the
cross-sections, is left for future work. 
Future work will also extend the approach to evaluate a probabilistic collision risk metric by integrating over volume intersections. 
Further, in our experiments, both methods are driven by a Gaussian tracker, so the check performed by ColNav-R is conditional on the $k\sigma$ sets containing
the true state. Pairing ColNav-R with a set-membership estimator is therefore a natural extension.







\balance
\bibliographystyle{IEEEtranBST/IEEEtran}
\bibliography{refs}

\end{document}